\documentclass{article}

\usepackage[preprint,nonatbib]{neurips_2022}

\usepackage{xspace}
\usepackage{wrapfig,lipsum,booktabs}
\usepackage{graphicx}
\usepackage[utf8]{inputenc} 
\usepackage[T1]{fontenc}    
\usepackage{hyperref}       
\usepackage{url}            
\usepackage{booktabs}       
\usepackage{amsfonts}       
\usepackage{nicefrac}       
\usepackage{microtype}      
\usepackage{xcolor}         
\usepackage{amsmath}
\usepackage{multirow}
\usepackage{dsfont}
\usepackage{amsmath}
\usepackage{amssymb}
\usepackage{booktabs}
\usepackage{amsmath}
\usepackage{multirow}
\usepackage{amsfonts}
\usepackage{amsthm}
\usepackage{multirow}
\usepackage{dsfont}
\newtheorem{theorem}{Theorem}[section]

\newtheorem{lemma}[theorem]{Lemma}
\theoremstyle{definition}
\newtheorem{definition}{Definition}[section]

\title{
Finding Diverse and Predictable Subgraphs for Graph Domain Generalization}
\author{Junchi Yu$^{1,2}$,
Jian Liang$^{1}$,
Ran He$^{1,2,3}$\thanks{Corresponding Author}\\
\and
$^1$NLPR\&CRIPAC Institute of Automation, Chinese Academy of Sciences, China\\
$^2$University of Chinese Academy of Sciences, China\\
$^3$Center for Excellence in Brain Science and Intelligence Technology, CAS, China\\
{\tt\small yujunchi2019@ia.ac.cn, jian.liang@lpr.ia.ac.cn, rhe@nlpr.ia.ac.cn}
}

\begin{document}

\maketitle
\begin{abstract}
This paper focuses on out-of-distribution generalization on graphs where performance drops due to the unseen distribution shift. 
Previous graph domain generalization works always resort to learning an invariant predictor among different source domains.
However, they assume sufficient source domains are available during training, posing huge challenges for realistic applications. 
By contrast, we propose a new graph domain generalization framework, dubbed as DPS, by constructing multiple populations from the source domains.
Specifically, DPS aims to discover multiple \textbf{D}iverse and \textbf{P}redictable \textbf{S}ubgraphs with a set of generators, namely, subgraphs are different from each other but all the them share the same semantics with the input graph.
These generated source domains are exploited to learn an \textit{equi-predictive} graph neural network (GNN) across domains, which is expected to generalize well to unseen target domains.
Generally, DPS is model-agnostic that can be incorporated with various GNN backbones.
Extensive experiments on both node-level and graph-level benchmarks shows that the proposed DPS achieves impressive performance for various graph domain generalization tasks.
{Code is attached in the supplementary.}
\end{abstract}

\section{Introduction}
Learning on graph-structured data is the fundamental problem in the machine learning field, ranging from many daily applications to scientific research \cite{fan2019graph,huang2022equivariant}. 
Recently, the Graph Neural Networks (GNNs) \cite{gcn} have become a \emph{de facto} standard in developing machine learning systems on graphs, which have shown superior performance on recommender system \cite{wu2020graph,li2019semi}, social network analysis \cite{bian2020rumor,fan2019graph}, biochemistry \cite{jin2018junction} and so on.
Despite the remarkable success, these models heavily rely on the I.I.D assumption that the training and testing data are independently drawn from an identical distribution \cite{li2022out}. 
When distribution shift incurs between the training and testing data, GNNs suffer from unsatisfactory performance, hindering their applications in real-world scenarios. 

Such distribution shift is attributed to the disparity between different domains (or environments) in the underlying data generation process. 
To generalize to unseen testing domains, recent advances in domain generalization attempt to learn an invariant predictor, which performs equally well on multiple source domains \cite{arjovsky2019invariant,krueger2021out}.
These methods usually assume access to abundant and diverse source domains for training \cite{matsuura2020domain, li2022out,qiao2020learning,wang2019learning}.
However, it may be impractically difficult to obtain sufficient source domains for real-world graphs \cite{li2022out}.
For example, in financial networks, we only have access to limited snapshots of the dynamic transaction graph due to the privacy issue.
A GNN trained with these snapshots can hardly exploit the invariance within finical behavior, hindering its applications on analysing future transactions. 
Besides, domain labels like age and gender are excluded to avoid fairness concerns during data collection \cite{friedler2019comparative}, making the final training set to be a single domain.
How to tackle the domain scarcity problem above poses a huge challenge for graph domain generalization.

To address this issue, a natural solution is to generate novel source domains by domain augmentation.
A pioneering method \cite{wu2022handling} learns several augmented domains by maximizing the variance of GNN's prediction loss.
However, the over-flexible domain augmentation strategy may produce implausible augmented domains, namely, some of learned augmentations are similar to each other \cite{wang2019learning,du2020learning}.
As a result, the insufficient augmented domains limits the performance gain in graph domain generalization, which motivates us to study the diversity during domain augmentation.

Since solely encouraging diversity leads to arbitrary semantics of augmented domains, we also expect the augmented domains to have consistent semantics with the source domain meanwhile.
To this end, we propose \textit{finding \textbf{D}iverse and \textbf{P}redictable \textbf{S}ubgraphs}, known as DPS, for graph domain generalization. 
Specifically, DPS consists of a set of generators and a GNN as the predictor.
Given the input graph from the source domain, DPS employs the generators to output diverse subgraphs which are predictable to the input graph label. 
To find predictable subgraphs, each generator is equipped with a variational distribution to minimize the risk of GNN's prediction. 
The subgraphs produced by different generators construct different augmented domains. 
To pursue diversity, we propose an energy-based regularization to enlarge the distances between the probability masses of different domains.
Thereby, the augmented domains are diverse and preserve consistent semantics to the source domain, avoiding from implausible augmentations.
Upon these augmented domains, the GNN is learnt to be equipredictive \footnote{By equipredictive, we mean that a predictor performs equally well on different domains, which is also known as the invariant predictor \cite{krueger2021out}.} across different domains, which is expected to generalize on unseen testing domains.
DPS is model-agnostic and can be adapted to both node-level and graph-level domain generalization tasks.
Extensive experiments demonstrate that DPS enjoys superior performance compared to existing algorithms on graph domain generalization.


To conclude, our contributions are in three-folds. First, we propose a new framework by constructing augmented domains with diverse and predictable subgraphs for graph domain generalization with scarce domains. Second, we propose a tractable subgraph generation method to efficiently find diverse and predictable subgrahs. Third, we conduct extensive experiments to validate the proposed DPS can be adapted to both graph-level and node-level tasks with different GNN backbones.


\section{Related Work}
\textbf{Graph Neural Networks.} The Graph Neural Network (GNN) has become a building-block for deep graph learning \cite{gcn}. It leverages the message-passing module to aggregate the adjacent information to the central node, which shows  expressive power on embedding rational data \cite{mmpn}. Various GNN variants have shown superior performance on social network analysis \cite{bian2020rumor}, recommender systerm \cite{wu2020graph}, physics \cite{huang2022equivariant, han2022equivariant} and  biochemistry \cite{jin2018junction, jin2020multi}. While GNNs have achieved notable success on many tasks, they heavily rely on the I.I.D assumption that the training and testing samples are drawn independently from the same distribution. This triggers concerns on the applications of GNN-based models in the real-world scenarios where there is a distribution shift between the training and testing data. Hence, it is imperative to investigate and improve the domain generalization ability of GNNs.



\textbf{Domain Generalization on Graphs.} Given the training samples from several source domains, domain generalization aims at generalizing deep models to unseen test domains \cite{wang2021generalizing}. To this end, researchers mainly resort to robust optimization \cite{qian2019robust,sagawa2019distributionally,hu2018does}, invariant representation/predictor learning \cite{krueger2021out,arjovsky2019invariant} and causal approaches \cite{buhlmann2020invariance,peters2016causal,heinze2018invariant}. Although domain generalization on Euclidean data has drawn much attention, seldom is there focus on its counterpart to the graph-structured data \cite{chen2022invariance}. Some pioneering works on graph domain adaptation focus on topology shift on synthetic and simple datasets \cite{baranwal2021graph,bevilacqua2021size}. Recently, researchers extend Out-of-distribution generalization methods to handle the distribution shift on graphs \cite{wu2022discovering,wu2022handling,li2022out,chen2022invariance}. However, the domain scarcity hinders from exploiting invariance across different source domains, which pose a huge challenge for learning an invariant GNN with generalization power. Noticeably, there are similar topics such as graph domain adaptation \cite{wu2020unsupervised,zhu2021shift} and graph transfer learning \cite{zhu2021transfer}. The main difference is that they usually assume access to part of test domains to adapt GNNs, while Graph Domain Generalization uses no samples from test domains.

\textbf{Subgraph Recognition.} The subgraph recognition problem aims at refining the graph structure for the improved performance in the graph representation learning \cite{yu2020graph}. Given an input graph, it generally removes the task-irrelevant edges, nodes or subgraphs and the classifier only takes the task-relevant part for prediction. For node classification tasks, it is popular to highlight the important neighborhood to learn robust and informative node representations via attention mechanism \cite{kim2022find,velickovic2017graph} and graph structure learning \cite{franceschi2019learning, yudonghan2020graph,zheng2020robust,chen2020iterative,sun2021graph}. At graph level, the information-theoretic approaches are employed to recognize a minimal sufficient subgraph which is free of noise and redundancy for the downstream tasks \cite{yu2020graph, yu2022improving}. Recently, it has drawn much attention to discover an invariant subgraph to empower GNN with out-of-distribution generalization ability \cite{chen2022invariance,wu2022discovering}. Our work differ from these works by exploiting diverse and predictable subgraphs as novel domains, which helps GNNs to generalize to unseen test domains.

\section{Method}
\subsection{Notations }
Let $\{(G_{i},Y_{i})|1\leq i\leq N\}$ be the training data from the source domain, which are sampled from the distribution $p(G,Y)=\sum_{e\in\mathcal{E}}p(G,Y|e)p(e)$. 
Here, $G\in \mathcal{G}$ is the graph with adjacent matrix $A$ and the node feature matrix $X$. $Y\in\mathcal{Y}$ is the ground-truth label of $G_{i}$. $e\in\mathcal{E}$ is the domain variable.
A graph neural network (GNN) $f:\mathcal{G}\rightarrow \mathcal{Y}$ maps $G$ to its label $Y$. 

\subsection{Main Idea \& Formulation}
When domain shift incurs between the source domain and the testing domain, GNN trained by Empirical Risk Minimization (ERM) \cite{vapnik1991principles} achieves unsatisfactory performance on the testing domain. This leads to the graph domain generalization, which aims to learn a generalized GNN to perform well on unseen domains. Recent studies on domain generalization or Out-of-distribution generalization generally learns an invariant predictor on multiple and diverse source domains \cite{krueger2021out,arjovsky2019invariant,sagawa2019distributionally}. However, it is difficult to obtain abundant domains for real-world graph applications \cite{li2022out, wu2022handling}, which poses a huge challenge for graph domain generalization.

To address such domain scarcity issue, we propose a new framework for graph domain generalization, namely \textbf{DPS}, which exploits diverse and predictable subgraphs from the training graphs. 
These subgraphs capture predictive information of the original graph on different aspects. Hence, they can construct multiple augmented domains to train GNNs for generalization.
Moreover, these subgraphs are sufficient to infer the original graph labels. This avoids from implausible domain augmentations, 
\begin{wrapfigure}{r}{0.7\textwidth}
\begin{center}
\centerline{\includegraphics[width=0.68\textwidth]{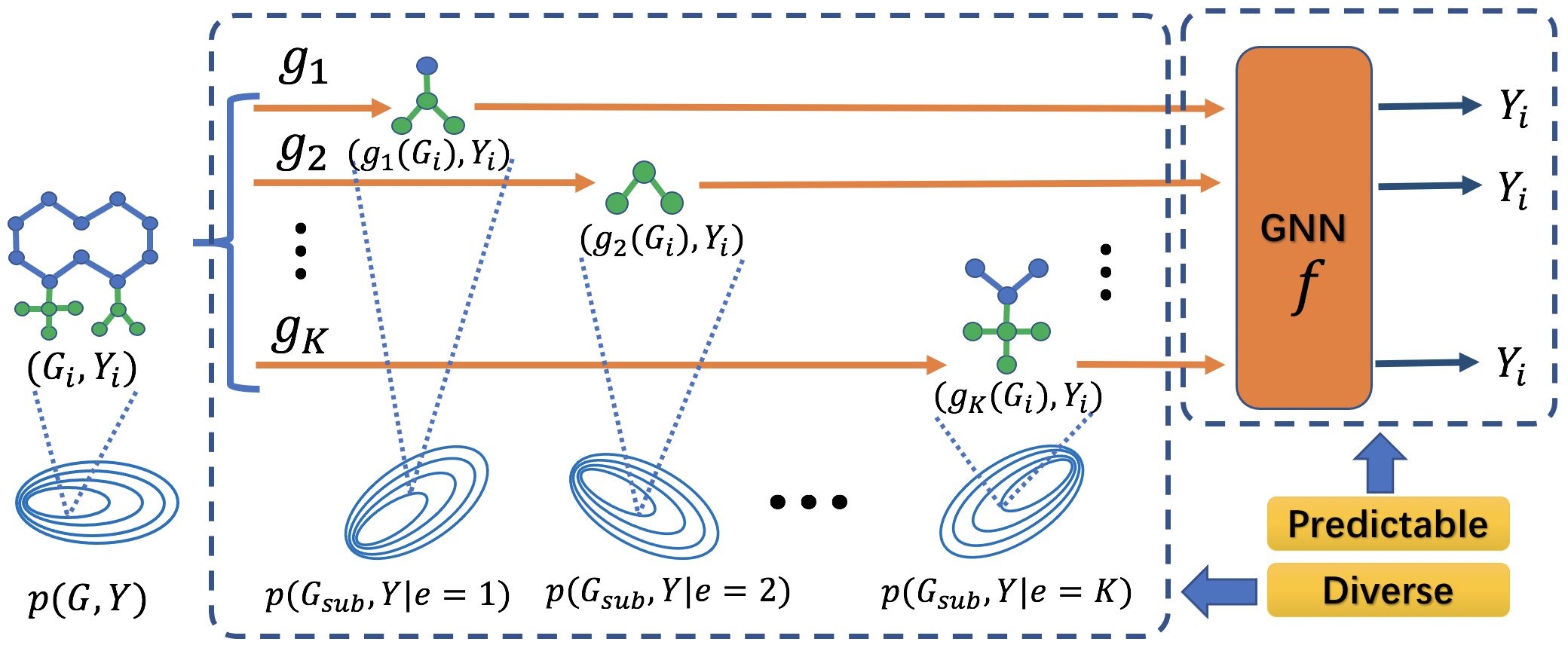}}
\end{center}
\vspace{-0.5cm}
\caption{The whole framework of DPS. DPS is equipped with $K$ subgraph generators $\{g_{i}\}_{i=1}^{K}$ and a GNN $f$ as the predictor. The subgraph generators find diverse and predictable subgraphs of the input graph to construct $K$ augmented domains. These domains share the same predictive distribution with the source domain, which helps GNN to be \textbf{equi-predictive} across the overall $K+1$ domains.}
\label{flowchart}
\vspace{-0.5cm}
\end{wrapfigure}
which lead to pessimistic predictor with degraded performance \cite{hu2018does,sagawa2019distributionally}. 
With the augmented domains, we can easily train an equipredictive GNN which performs equally well on different domains, even with standard ERM. To elaborate how DPS works, we start with the definition of predictable subgraph generator. 

\theoremstyle{definition}
\begin{definition}[Predictable Subgraph Generator]
For graph $G$ with label $Y$, the subgraph generator $G_{sub}=g(G)$ is said to be predictable subgraph generator if $g=\arg\max_{g}I(G_{sub},Y)$.
\end{definition}

Here $I(\cdot)$ is Shannon Mutual Information. $g(\cdot)$ represents a deterministic or stochastic subgraph generator and the produced $G_{sub}$ is called the \textit{predictable subgraph}. Maximizing $I(G_{sub},Y)$ leads to $I(G_{sub},Y)=I(G,Y)$ due to the Data Processing Inequality. Hence, $G_{sub}$ is also \textit{sufficient} to predict $Y$ \cite{yu2021recognizing}. 
Noticeably, the predictable subgraph is not unique due to the redundancy of graph-structured data \cite{conf/icml/FranceschiNPH19,jin2020graph-structure,yu2021recognizing}. 
For example, for a mutagenic molecule, the subgraphs which have the functional groups (e.g. aromatic $-NO_{2}$ or $-NH_{2}$) are all sufficient to cause the mutagenic effect \cite{kawai1987mutagenicity,helma2006lazy,gnnexplainer}. 
Hence, these subgraphs are all predictable subgraphs with diverse topology.
This motivates us to mine diverse and predictable subgraphs from the training graphs to mimic different augmented domains.
\begin{definition}[Diverse and Predicable Subgraphs] 
\label{diverse-predictable}
$G_{sub}^{e=1}=g_{1}(G)$ and $G_{sub}^{e=2}=g_{2}(G)$ are two diverse and predicable subgraphs, if: $g_{1}(\cdot)$ and $g_{2}(\cdot)$ are two predictable subgraph generators, and $g_{1}(\cdot) \neq g_{2}(\cdot)$.
\end{definition}

Suppose there are $K$ subgraph generators $\{g_{i}\}_{i=1}^{K}$ to produce diverse and predicable subgraphs. We can treat these subgraphs as samples from $K$ different domains, leading to
$p(G_{sub},Y)=\sum_{i=1}^{K}p(G_{sub},Y|e=i)p(e=i)$. Here $p(G_{sub},Y|e=i)$ is the $i$-th domain induced by $g_{i}$. 
The labels of samples in $K$ augmented domains are well defined since these subgraphs are all predictable. Thus, these augmented domains are diverse and share consistent semantic with the source domain. More importantly, these $K$ domains share the same predictive relationship $p(Y|e,G_{sub})=p(Y|G_{sub})$. 
Formally, we have the following lemma.


\begin{lemma}
Given the training data $(G,Y)\in p(G,Y)$, and $K$ different predictable subgraph generators $\{g_{i}\}_{i=1}^{K}$. The obtained 
$K$ domains constructed by the diverse predictive subgraphs have an invariant predictive distribution: $p(Y|e,G_{sub})=p(Y|G_{sub})$.
\label{lemma}
\end{lemma}

The proof of Lemma~\ref{lemma} is in Appendix. Lemma~\ref{lemma} indicates that the predictive distribution is invariant across $K$ augmented domains. Such property is equal to the sufficiency criterion for learning a invariant predictor \cite{federici2021information}. In most domain generalization tasks, the predictive distribution $p(Y|e,G_{sub})$ varies across different domains. Thus, prior works usually employ a regularization to learn a invariant predictor \cite{krueger2021out,arjovsky2019invariant}. Differently, we generate augmented domains embedded with an invariant relationship between the inputs and their labels. 
A GNN $f$ can approaches the invariant predictive distribution $p(Y|G_{sub})$ by minimizing the empirical risk on $K$ domains. This guides $f$ to be equipredictive on different domains. 


\begin{theorem}
With $K$ domains generated by DPS, a GNN $f$ can approach an invariant predictor by empirical risk minimization.
\end{theorem}


For simplicity, we denote the source training domain $p(G,Y)$ as the (K+1)-th domain, i.e. $p(G_{sub},Y|e=K+1)$, to form $K+1$ source domains. 
With all $K+1$ domains for training, we aim to learn a GNN, denoted as $f$, which is expected to generalize to unseen test domains. Thus, the whole objective is as follows:
\begin{equation}
\begin{aligned}
&\min_{f} \frac{1}{K+1} \sum_{i=1}^{K+1}\mathcal{L}_{f}(G_{sub},Y|e=i)\\
&s.t. \; G_{sub}=g(G), g\in\{g_{i}\}_{i=1}^{K}.
\end{aligned}
\label{og-obj}
\end{equation}
Although it is promising to enrich the source domain and learn an equipredictive GNN with DPS, it is challenging to discover these diverse and predictive subgraphs due to the discrete and exponentially large subgraph space.

\subsection{Finding Diverse and Predictable Subgraphs for Generalization}
To address these issues, we propose a learning-based method to to efficiently discover diverse and predictable subgraphs. Specifically, we parameterize the subgraph generators in Eqn.~\ref{og-obj} with GNNs. Then, the subgraph generation process can be viewed as the compression of input graphs, since $G_{sub}$ only preserves a part of topological information of the input graph $G$.
This is formulated as the following objective:
\begin{equation}
\begin{aligned}
\min_{g} \frac{1}{K}\sum_{i=1}^{K}I(G, G_{sub}|e=i).
\end{aligned}
\label{compression}
\end{equation}
Here $I(\cdot)$ denotes the mutual information (MI). 
Unfortunately, Eqn.~\ref{compression}  is notoriously difficult to optimize since MI is intractable to compute. Estimating the MI with estimators such as Donsker-Varadhan representation \cite{mine} is unstable and computationally expensive for graph-structured data, and usually lead to degraded results \cite{yu2021recognizing,yu2022improving}.
Hence, one must specify appropriate subgraph generators $\{g_{i}\}_{i=1}^{K}$ for a tractable form of Eqn.~\ref{compression}. To this end, we leverage the node sampling process for subgraph generation. 
For node $v\in G$, $g_{i}$ sample a node mask $m_{v,g_{i}}\sim \mathrm{Bernoulli}(p_{v,g_{i}})$, where  $p_{v,g_{i}}$ is a learnable probability. Denote the mask matrix of all nodes as $M_{g_{i}}$, we obtain the subgraph $G_{sub}$ by applying $M_{g_{i}}$ to all the nodes in $G$: $G_{sub}=G\odot M_{g_{i}}$
Specifically, given a graph $G=\{A,X\}$, $g_{i}$ employs a $l$-layer GNN and a Multi-Layer Perceptron (MLP) to output $p_{v,\mathcal{T}_{i}}$:
\begin{equation}
\begin{aligned}
H=\mathrm{GNN}(A,X),\ 
p_{v,g_{i}}=\mathrm{Sigmoid}(\mathrm{MLP}(h_{v})).
\end{aligned}
\label{probability}
\end{equation}
Here, $H$ is the node embedding matrix and $h_{v}$ is the embedding of node $v$. The output of MLP is mapped into [0,1] via the Sigmoid function. 
Since the node sampling process in non-differentiable, we further employ the concrete relaxation \cite{jang2016categorical,gal2017concrete} for $m_{v,\mathcal{T}_{i}}$:
\begin{equation}
\begin{aligned}
\hat{m}_{v,g_{i}} = \mathrm{Sigmoid}(\frac{1}{t}\log{\frac{p_{v,g_{i}}}{1-p_{v,g_{i}}}}+\log{\frac{u}{1-u}}),
\end{aligned}
\end{equation}
where $t$ is the temperature parameter and $u\sim \mathrm{Uniform}(0,1)$.
%
%
Then, we seek for a tractable upper bound of the objective in Eqn.~\ref{compression}.
In fact, we have the following inequality:
\begin{equation}
\begin{aligned}
\mathrm{I}(G,G_{sub}|e=i)&=\mathbb{E}_{G,G_{sub}} \log{\frac{p(G_{sub}|G,e=i)}{q(G_{sub}|e=i)}}-\mathrm{KL}[p(G_{sub}|e=i)|q(G_{sub}|e=i)]\\
&\leq \mathbb{E}_{G\sim p(G)} \mathrm{KL}[p(G_{sub}|G,e=i)|G)|q(G_{sub}|e=i)].
\end{aligned}
\label{inequality}
\end{equation}
Here, $\mathrm{KL}$ is the KL-divergence. The inequality is due to the fact that KL-divergence is non-negative. 
The posterior $p(G_{sub}|G,e=i)|G)$ in Eqn.~\ref{inequality} is parameterized by $g_{i}(\cdot)$,
which can be factorized into the multiplication of node sampling probabilities $\prod_{v\in G} \mathrm{Bernoulli}(p_{v,g_{i}})$.
The specification of the prior $p(G_{sub}|e=i)$ in Eqn.~\ref{inequality} is chosen to be the non-informative distribution $\prod_{v\in G}\mathrm{Bernoulli}(\alpha)$ following \cite{alemi2016deep,wu2020gib}. We set $\alpha=0.5$, which encodes equal probability of sampling or dropping nodes in prior knowledge.
%
%
Thus, we finally reach a tractable upper bound of Eqn.~\ref{compression}:
\begin{equation}
\begin{aligned}
\mathcal{L}_{kld}(g)&= \frac{1}{K}\sum_{i=1}^{K}
\mathbb{E}_{G\sim p(G)} \mathrm{KL}[p(G_{sub}|G,e=i)|q(G_{sub|e=i})]\\
&=\frac{1}{NK}\sum_{i}^{K}\sum_{j}^{N}\sum_{v\in G_{j}}\mathrm{KL}[\mathrm{Bernoulli}(p_{v,g_{i}})|\mathrm{Bernoulli}(\alpha)].
\end{aligned}
\label{kld}
\end{equation}

\textbf{Enforcing Diverse and Predictable Subgraphs.} After obtaining subgraphs with the subgraph generators, we first constrain these subgraphs to be predictable. By definition, the predictable subgraphs are maximally predictive to input graph labels. This condition can be converted into minimize the following cross-entropy loss:
\begin{equation}
\begin{aligned}
\mathcal{L}_{CE}(f,\mathcal{T}) &= \frac{1}{N(K+1)}\sum_{i=1}^{K+1}
\sum_{j=1}^{N}\{-\log{f(g_{i}(G_{j}))}[Y]\}.
\end{aligned}
\label{predictable}
\end{equation}
where $f$ is the GNN predictor in Eqn.~\ref{og-obj}. $f(\cdot)[Y]$ denotes the $Y$-th logit output by $f$. Eqn.~\ref{predictable} plays two roles. First, it encourages different $g_{i}$ generate predictable subgraphs for augmentation. Second, it minimizes the empirical risk of $f$ across all domains, which guides $f$ to make correct prediction. 

To impose diversity across the subgraphs in different domains, we need to compute pair-wise distances between distributions of two domains.
Although the Erd{\H{o}}s-R{\'e}nyi graph model specifies the probability density of graphs generated from specific distribution \cite{erdHos1960evolution}, estimating the density of arbitrary graphs is still difficult for their irregular and discrete nature \cite{lecun2006tutorial}. The brutal force method is to go through all the samples from the underlying distribution of graphs, which is computationally prohibited.



\textbf{Energy-based Modeling of Subgraph Density.}
To this end, we introduce an energy-based model (EBM) to specify the probability density of subgraph distribution:
$p(G_{sub}|e=i) \propto \exp{-E_{\theta}(G_{sub}|e=i)}$.
Here $E_{\theta}:\mathcal{G}\rightarrow \mathcal{R}$ is the energy score. 
Interestingly, $E_{\theta}$ can be derived from the predictor $f$ in Eqn.~\ref{predictable} as follows \cite{grathwohl2019your}:
\begin{equation}
\begin{aligned}
E_{\theta}(G_{sub}|e=i)=-\log\sum_{Y\in\mathcal{Y}}\exp{f(g_{i}(G))[Y]}.
\end{aligned}
\label{energy}
\end{equation}
Since the subgraph density is proportion to the energy score induced by the classification model, we propose an energy-based regularization:
\begin{equation}
\begin{aligned}
\mathcal{L}_{dist}(f,g) = \frac{1}{N}\sum_{i=1}^{N} \frac{2}{K(K+1)} \sum_{j=1}^{K}\sum_{k=j+1}^{K+1} \frac{1}{2}[E_{f}(g_{j}(G_{i})) - E_{f}(g_{k}(G_{i}))]^{2}
\end{aligned}
\label{dist-loss}
\end{equation}

Please refer to Appendix for more details on Eqn.~\ref{energy} and Eqn.~\ref{dist-loss}.
%
%
%
%
%
%
The energy score in Eqn.~\ref{dist-loss} is scalable to measure the pair-wise distance between the two augmented domains. Notice $\mathcal{L}_{dist}(f,g)$ is unbounded and can be arbitrary large, which leads to unstable training process. To avoid this scenario, we let $f$ and $g$ play a minmax game where $g$ maximizes $\mathcal{L}_{dist}(f,g)$ and $f$ minimizes $\mathcal{L}_{dist}(f,g)$. Combine the objective in Eqn.~\ref{predictable}, Eqn.~\ref{kld} and Eqn.~\ref{dist-loss}, we obtain the total loss function:
\begin{equation}
\begin{aligned}
&\min_{f}\mathcal{L}_{CE}(f,g^{*}) + \alpha\mathcal{L}_{dist}(f,g^{*})\\
&s.t. g^{*} = \arg\min_{g} \mathcal{L}_{CE}(f,g) +\beta\mathcal{L}_{kld}(g) - \alpha\mathcal{L}_{dist}(f,g).
\end{aligned}
\label{total-loss}
\end{equation}
Here $\alpha$ and $\beta$ are hyper-parameters.
The optimization problem in Eqn.~\ref{total-loss} is a bi-level optimization problem \cite{yu2020graph}. In practice, we first obtain a sub-optimal $g^{*}$ by by optimizing $g$ for $T$ steps in the inner loop. Then, we use the sub-optimal $g^{*}$ as a proxy in the outer loop to optimize $f$. We provide pseudo code for optimizing Eqn.~\ref{total-loss} in Appendix.

\subsection{Extension to Node-level Tasks}
\label{node-level-task}
We proceed to introduce the extension of DPS on node-level tasks.
Different from the graph classification task,  the nodes in one graph are treated from the same domain in node classification task. Moreover, these nodes are associated with their neighborhoods with the egdes, leading to non-independence in the training samples. Hence, we follow the prior works by taking a local view of the nodes and relate them with K-hop ego-graphs \cite{wu2022handling,zhu2021transfer}. For example, $N_{i}$ is associated with its 1-hop ego-graph $G_{i}=(A_{i}, X_{i})$, where $A_{i}$ is the adjacent matrix of the 1-hop subgraph centered at $N_{i}$ and $X_{i}$ is the neighborhood node feature matrix. 
Therefore, it allows to discover diverse and predictable subgraph hierarchically and is compatible with the messaging-passing procedure in GNNs.
Moreover, instead of generating subgraph via node sampling, we employ a edge sampling procedure for node classification task. Since the nodes are treated as training samples, we can not fully utilize the training samples by directly dropping nodes. Instead, for a central node, we can block its undesired neighbors from the message-passing and aggregation process by dropping their edges, which is equivalent to node sampling in 1-hop ego-graphs. Hence, for the edge $e_{uv}$ between node $u$ and $v$, the subgraph generator $g_{i}$ is equipped with a learnable probability $p_{uv,g_{i}}$. Then the edge mask is sampled from $m_{uv,g_{i}}\sim \mathrm{Bernoulli}(p_{uv,g_{i}})$. Hence, for node classification task, the loss in Eqn.~\ref{kld} is as follows:
\begin{equation}
\begin{aligned}
\mathcal{L}_{kld}(g)
&=\frac{1}{N(K+1)}\sum_{i=1}^{K+1}\sum_{j=1}^{N}\sum_{uv\in \mathcal{E}}\mathrm{KL}[\mathrm{Bernoulli}(p_{uv,g_{i}})|\mathrm{Bernoulli}(\alpha)].
\end{aligned}
\label{kld-edge}
\end{equation}
\begin{table}[t]
  \centering
  \setlength{\tabcolsep}{1.0mm}{
  \footnotesize
  \caption{Performances of different methods on OOD graph classification tasks. We report mean and standard deviation of Accuracy. In Spurious-Motif dataset, $b$ is the indicator of spurious correlation.}
    \begin{tabular}{c|cccc|c|c}
    \toprule
    \multirow{2}[4]{*}{Method} & \multicolumn{4}{c|}{Spurious-Motif} & \multirow{2}[4]{*}{MUTAG} & \multirow{2}[4]{*}{D\&D} \\
\cmidrule{2-5}          & b=0.33 & b=0.5 & b=0.7 & b=0.9 &       &  \\
    \midrule
    ERM \cite{vapnik1991principles}   & 34.93 $\pm$ 0.75 & 34.36 $\pm$ 2.91 & 34.73 $\pm$ 0.21 & 32.33 $\pm$ 0.24 & 70.11 $\pm$ 3.23 & 53.74 $\pm$ 4.47 \\
    V-Rex \cite{krueger2021out}& 40.24 $\pm$ 1.68 & 39.70 $\pm$ 2.63 & 39.06 $\pm$ 2.64 & 38.14 $\pm$ 3.30 & 70.49 $\pm$ 4.73 & 73.03 $\pm$ 3.05 \\
    IRM \cite{arjovsky2019invariant}  & 40.91 $\pm$ 5.34 & 40.64 $\pm$ 2.57 & 40.07 $\pm$ 1.60 & 37.15 $\pm$ 2.08 & 69.74 $\pm$ 3.35 & 73.27 $\pm$ 1.75 \\
    \midrule
    Attention \cite{conf/nips/KnyazevTA19} & 35.68 $\pm$ 0.73 & 34.42 $\pm$ 0.34 & 33.72 $\pm$ 0.29 & 33.97 $\pm$ 0.34 & 68.23 $\pm$ 3.12 & 52.91 $\pm$ 5.36 \\
    Top-k pool \cite{gao2019graph} & 34.45 $\pm$ 0.44 & 33.87 $\pm$ 0.15 & 33.58 $\pm$ 0.25 & 34.10 $\pm$ 0.18 & 72.36 $\pm$ 2.65 & 66.33 $\pm$ 3.17 \\
    GIB \cite{yu2020graph}   & 39.71 $\pm$ 3.91 & 37.45 $\pm$ 3.67 & 36.43 $\pm$ 3.61 & 35.42 $\pm$ 0.83 & 56.83 $\pm$ 3.59 & 54.34 $\pm$ 7.18 \\
    VGIB \cite{yu2022improving}  & 37.33 $\pm$ 0.78 & 37.84 $\pm$ 0.57 & 34.97 $\pm$ 1.05 & 34.69 $\pm$ 0.90 & 81.38 $\pm$ 2.31 & 65.29 $\pm$ 6.43 \\
    DIR \cite{wu2022discovering}  & 46.87 $\pm$ 2.52 & 43.30 $\pm$ 3.07 & 43.84 $\pm$ 2.13 & 38.65 $\pm$ 1.19 & 83.53 $\pm$ 4.17 & 73.17 $\pm$ 7.43 \\
    \midrule
    DPS  & \textbf{51.91 $\pm$ 4.23} & \textbf{46.63 $\pm$ 5.04} & \textbf{47.23 $\pm$ 4.20} & \textbf{44.11 $\pm$ 2.19} & \textbf{86.06 $\pm$ 1.50} & \textbf{75.31 $\pm$ 8.17} \\
    \bottomrule
    \end{tabular}%
  \label{graph-level}}%
\end{table}%

\section{Experiments}
\label{section-experiment}
In this section, we extensively evaluate the proposed DPS on both node-level and graph-level tasks with different types of distribution shift.

\subsection{Out-of-distribution Graph Classification}
We first evaluate DPS on out-of-distribution (OOD) graph classification. We train a GNN on a single source domain and evaluate its performance on unseen testing domains.

\textbf{Datasets.} We employ Spurious-Motif \cite{gnnexplainer}, MUTAG \cite{nr} and D\&D \cite{conf/nips/KnyazevTA19} datasets for OOD graph classification. The Spurious-Motif dataset consists of 18000 synthetic graphs. Each graph is generated by attaching one base (Tree, Ladder, Wheel, denoted as $S=0,1,2$) to a motif (Cycle, House, Crane, denoted as $C=0,1,2$). The graph label $Y$ is consistent to the class of motif. For the training graphs, the base is chosen with probability $P(S)=b\times\mathds{1}(S=C) + \frac{1-b}{2}\times\mathds{1}(S\neq C)$ to create spurious correlation. We change $b$ to impose different bias on the training graphs. For testing graphs, the motifs and bases are randomly connected. For D\&D and MUTAG datasets, we choose the graph size as the shift \cite{conf/nips/KnyazevTA19,li2022out}. Specifically, we choose the graphs in D\&D dataset with less than 200 nodes for training, those with 200-300 nodes for validation, and graphs larger than 300 nodes for testing. For MUTAG, we select graphs with less than 15 nodes for training, those with 15-20 nodes for validation, and graphs larger than 20 nodes for testing. We report accuracy (Acc) for these datasets.

\begin{table}[t]
  \centering
  \setlength{\tabcolsep}{1.0mm}{
  \footnotesize
  \caption{Test ROC-AUC on Twitch-Explicit dataset. Each method is trained on single source domain. For a fair comparision, both EERM and DPS generate 3 augmented domains.}
    \begin{tabular}{c|c|ccccc}
    \toprule
    Backbone & Method & ES    & FR    & PTBR  & RU    & TW \\
    \midrule
    \multirow{3}[2]{*}{GCN \cite{gcn}} & ERM \cite{vapnik1991principles}   & 52.50 $\pm$ 4.09 & 54.92 $\pm$ 2.60 & 48.78 $\pm$ 7.45 & 50.49 $\pm$ 1.82 & 48.95 $\pm$ 2.31 \\
          & EERM \cite{wu2022handling}  & 54.17 $\pm$ 5.04 & 54.10 $\pm$ 1.76 & 49.49 $\pm$ 7.96 & 51.34 $\pm$ 1.67 & 49.83 $\pm$ 3.15 \\
          & DPS  & \textbf{57.97 $\pm$ 2.96} & \textbf{55.87 $\pm$ 2.66} & \textbf{59.96 $\pm$ 2.12} & \textbf{52.73 $\pm$ 0.67} & \textbf{52.60 $\pm$ 2.64} \\
    \midrule
    \multirow{3}[2]{*}{GraphSAGE \cite{conf/nips/HamiltonYL17}} & ERM \cite{vapnik1991principles} & 66.73 $\pm$ 0.32 & 62.00 $\pm$ 0.26 & 65.13 $\pm$ 0.66 & 56.04 $\pm$ 0.19 & 59.23 $\pm$ 0.52 \\
          & EERM \cite{wu2022handling} & 66.79 $\pm$ 0.19 & 61.89 $\pm$ 0.39 & 65.08 $\pm$ 0.21 & 56.26 $\pm$ 0.12 & 59.49 $\pm$ 0.31 \\
          & DPS  & \textbf{66.86 $\pm$ 0.35} & \textbf{62.21 $\pm$ 0.36} & \textbf{65.22 $\pm$ 0.65} & \textbf{56.70 $\pm$ 0.42} & \textbf{59.70 $\pm$ 0.84} \\
    \bottomrule
    \end{tabular}}%
  \label{twitter}%
\end{table}%
\begin{figure}
\centering
\includegraphics[width=\textwidth]{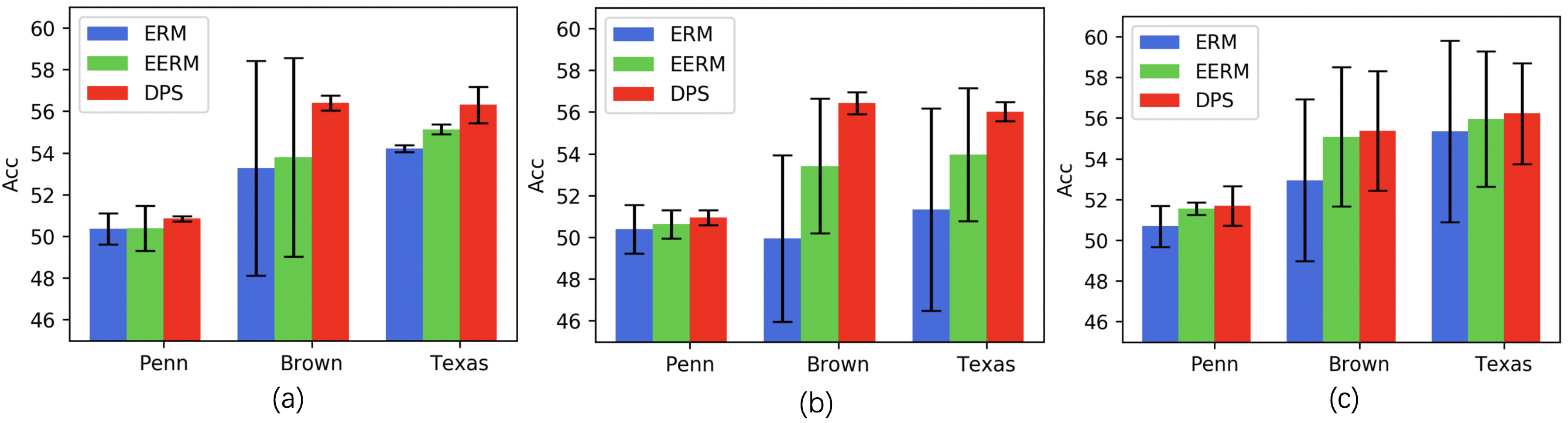}
\centering
\caption{We report performances of different methods with 3 source domain combinations on Facebook-100 dataset: (a). John Hopkins + Caltech + Amherst; (b).Bingham + Duke + Princeton; and (c). WashU + Brandeis+ Carnegie. We report mean and standard deviation of Accuracy across different runs.}
\label{facebook}
\end{figure}

\begin{table}[t]
  \centering
  \footnotesize
  \setlength{\tabcolsep}{1.5mm}{
  \caption{Mean and standard deviation of Accuracy (Acc) on OGB-Arxiv dataset.}
    \begin{tabular}{c|c|c|c|c|c|c}
    \toprule
    Test Domain & \multicolumn{2}{c|}{14-16} & \multicolumn{2}{c|}{16-18} & \multicolumn{2}{c}{18-20} \\
    \midrule
    Backbone & \multicolumn{1}{c|}{APPNP \cite{klicpera2018predict}} & \multicolumn{1}{c|}{SGGCN \cite{wu2019simplifying}
    } & \multicolumn{1}{c|}{APPNP} & \multicolumn{1}{c|}{SGGCN} & \multicolumn{1}{c|}{APPNP} & \multicolumn{1}{c}{SGGCN} \\
    \midrule
    ERM \cite{vapnik1991principles}  & 46.30 $\pm$ 0.35 & 40.52 $\pm$ 1.24 & 43.75 $\pm$ 0.40 & 38.23 $\pm$ 2.15 & 39.78 $\pm$ 0.41 & 34.62 $\pm$ 2.14 \\
    EERM \cite{wu2022handling}  & 46.42 $\pm$ 0.46 & 42.37 $\pm$  2.37 & 44.53 $\pm$ 0.54 & 39.91 $\pm$ 2.07 & \textbf{43.24 $\pm$ 0.79} & 37.73 $\pm$ 1.42 \\
    DPS   & \textbf{47.66 $\pm$ 0.24} & \textbf{44.32 $\pm$ 0.47} & \textbf{45.09 $\pm$ 0.29} & \textbf{41.95 $\pm$ 0.60} & 41.22 $\pm$ 0.25 & \textbf{38.89 $\pm$ 0.71} \\
    \bottomrule
    \end{tabular}%
  \label{arxiv}}%
\end{table}%

\begin{figure}[t]
\begin{center}
\centerline{\includegraphics[width=1.0\textwidth]{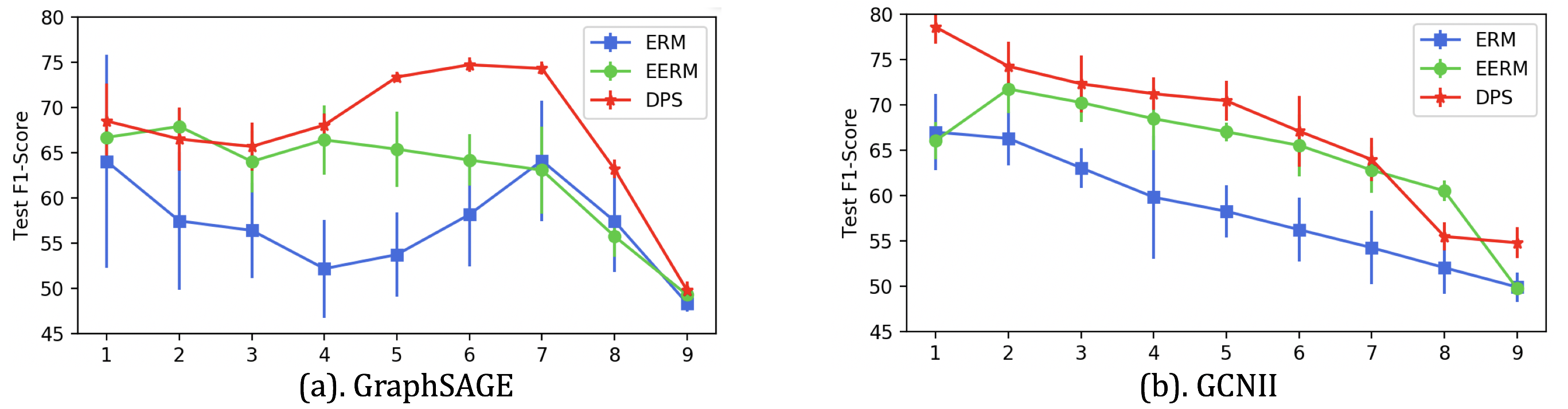}}
\end{center}
\vspace{-0.7cm}
\caption{Test F1-Score on 9 testing folds in ELLIPTIC dataset.}
\label{elliptic}
\vspace{-0.7cm}
\end{figure}
\textbf{Baselines.} We first compare our method with invariant learning methods, including V-Rex \cite{krueger2021out} and IRM \cite{arjovsky2019invariant}.
These methods aims to learn a invariant predictor/representation for OOD generalization. Since there is only one domain for training, we randomly group graphs to mimic different domains to instantiate V-Rex and IRM. Moreover, we compare DPS with various interpretable baselines, such as Attention-based Pooling \cite{conf/nips/KnyazevTA19}, TopK-Pooling \cite{gao2019graph}, GIB \cite{yu2020graph}, VGIB \cite{yu2022improving}, DIR \cite{wu2022discovering}. These methods highlight an important or invariant subgraph for prediction. We evaluate the model with highest validation accuracy and report mean and standard deviation of 10 runs for each method.

\textbf{Performance.} As shown in Table~\ref{graph-level}, DPS outperforms the baseline methods on both synthetic and real-world datasets, with up to 5\% absolute performance gain. IRM and V-Rex exceed standard ERM on Spurious-Motif and D\&D, but only achieve comparable performance on MUTAG. Hence, the performances of invariant learning methods with random domain partition are sometimes subject to the datasets. For the interpretable methods except for DIR, they only surpass ERM when there is no spurious correlation ($d=0.33$). The reason is that they find important subgraphs rather than invariant subgraphs, leading to the unsatisfying performance on OOD graph classification. Noticeably, DIR consistently outperforms ERM and invariant learning methods, indicating that incorporating invariant learning and graph topology can significantly facilitate OOD graph classification. 

\subsection{Node-level Domain Generalization}
We proceed apply DPS to node-level domain generalization where the training nodes are from one or several limited domains. Then, we evaluate the predictive performance of DPS on nodes from unseen testing domains. 
We consider two types of distribution shift: the spatial shift and temporal shift. 

\textbf{Datasets \& Metrics.} For the spatial shift, we adopt Twitch-Explicit \cite{rozemberczki2021multi} and Facebook-100 \cite{traud2012social} datasets for evaluation. These datasets contains different social networks which are related to different locations such as campus and districts. For example, Twitch-Explicit contains seven social networks, including DE, ENGB, ES, FR, PTBR, RU and TW. Following the protocol in prior work \cite{wu2022handling}, we employ DE for training, ENGB for validation and the rest five network for testing. For Facebook-100 dataset, we choose different combinations of three graphs for training, two for validation and the rest three graphs for testing. We report ROC-AUC and Accuracy (Acc) for Twitch-Explicit and Facebook-100 respectively.
For the temporal shift, we use a citation network OGB-Arxiv \cite{hu2020open} and a dynamic financial dataset ELLIPTIC \cite{pareja2020evolvegcn}. For OGB-Arxiv, we employ the papers published before 2011 for training, from 2011$\sim$2014 for validation and those within 2014$\sim$2016/2016$\sim$2018/2018$\sim$2020 for testing. For ELLIPTIC, we split the whole dataset in to different snapshots, and use 5/5/33 for training, validation and testing. The testing domains are further chronologically clustered into 9 folders for the convenience of comparing the performances of different methods. We report Test F1 Score and Accuracy (Acc) for ELLIPTIC and OGB-Arxiv respectively.

\textbf{Baselines.} We compare the performance of the proposed DPS with Empirical Risk Minimization (ERM) and the state-of-the-art node generalization method, Explore-to-Extrapolate Risk Minimization (EERM) \cite{wu2022handling}. For a fair comparison, we generate 3 augmented domains for EERM and DSP. 
We further plug different methods into various GNN backbones, such as GCN \cite{gcn}, GraphSAGE \cite{conf/nips/HamiltonYL17}, APPNP \cite{klicpera2018predict}, SGGCN \cite{wu2019simplifying} and GCNII \cite{chen2020simple}, to extensive evaluate their performance. 
We evaluate the model with highest validation accuracy and report mean and standard deviation of 10 runs for each method.

\textbf{Performance.} We report the results on Twitch-Explicit in Table~\ref{twitter}. The proposed DSP exceed the baselines on most testing domains. Since there is only one source domain for training, ERM is difficult to generalize to unseen testing domains with unsatisfactory performances. The recently proposed EERM employs an extrapolation-based paradigm for domain augmentation. We find that EERM can sometimes under-perform the standard ERM due to the implausible augmentations which are insufficient to offer GNN with OOD knowledge. In Figure~\ref{facebook}, we compare different methods with GCN backbone on Facebook-100 dataset. We can see that DPS exceeds the baselines with different source domain combinations. Moreover, DPS shows relatively small variance in performance across 10 runs, which shows the stability of DPS.

For the temporal shift on nodes, we first plug different methods into APPNP and SGGCN backbones and evaluate their performances on OGB-Arxiv dataset. As shown in Table~\ref{arxiv}, DPS outperforms the baselines in five cases out of six with stable results in different runs. Then, we report the results on Elliptic dataset in Figure~\ref{elliptic}. DPS outperforms the baseline methods up to 10\% absolute performance gain. Moreover, we observe that performance of DSP can vary when we adopt different backbones. 
When we adopt GraphSAGE as the backbone, DPS shows an increasing performance between 3-7 folds. Such phenomenon can also be observed in the results of ERM. For GCNII, the accuracy of DPS declines slowly between 3-7 folds. The main difference in backbone is that GraphSAGE employs a random sampling procedure in message-passing, which may be beneficial to graph domain generalization, as shown in the performance of DPS-Random in Table~\ref{twitter}. Thus, different GNN backbones can influence the generalization performance of different methods. 
\begin{table}[t]
  \centering
  \footnotesize
  \caption{Ablation study on Twitch-Explicit dataset. DPS surpasses all variant models.}
    \begin{tabular}{c|ccccc}
    \toprule
    Method & ES    & FR    & PTBR  & RU    & TW \\
    \midrule
    DPS  & \textbf{57.97 $\pm$ 2.96} & \textbf{55.87 $\pm$ 2.66} & \textbf{59.96 $\pm$ 2.12} & \textbf{52.73 $\pm$ 0.67} & 52.60 $\pm$ 2.64 \\
    DPS-Random & 54.64 $\pm$ 3.73 & 52.54 $\pm$ 2.60 & 55.74 $\pm$ 4.44 & 49.65 $\pm$ 1.12 & 49.85 $\pm$ 4.20 \\
    DPS-Rex & 57.75 $\pm$ 3.75 & 53.77 $\pm$ 0.84 & 55.40 $\pm$ 9.04 & 52.47 $\pm$ 0.39 & \textbf{54.66 $\pm$ 0.53} \\
    DPS w/o $\mathcal{L}_{dist}$ & 57.28 $\pm$ 3.49 & 54.80 $\pm$ 1.37 & 57.73 $\pm$ 7.23 & 52.55 $\pm$ 0.93 & 52.67 $\pm$ 2.21 \\
    DPS w/o $\mathcal{L}_{kld}$ & 55.81 $\pm$ 2.21 & 54.94 $\pm$ 2.49 & 57.49 $\pm$ 2.17 & 51.76 $\pm$ 0.91 & 50.71 $\pm$ 2.47 \\
    \bottomrule
    \end{tabular}%
    \vspace{-0.5cm}
  \label{ablation}%
\end{table}%
\subsection{Discussions}
We derive two variant model of DPS, namely DPS-Random and DPS-Rex. DPS-Random randomly drops a portion of edges in the training graph \cite{rong2020dropedge}. DPS-Rex replaces the energy-regularization in Eqn.~\ref{total-loss} with the variance of loss in different domains. Moreover, we remove $\mathcal{L}_{dist}$ and $\mathcal{L}_{kld}$ in Eqn.~\ref{total-loss} to study their effects on the performance. The performances are shown in Table~\ref{ablation}. Interestingly, DPS-Random achieves competitive performance by randomly dropping edges, which indicates that network sparsification may be a promising solution to graph domain generalization. We leave in-depth discussion in the future work. DPS-Rex underperforms DPS on five out of six cases. The reason is that maximizing the loss variance is insufficient to generate diverse augmentations.
\begin{wraptable}[11]{r}{0.50\textwidth}
  \centering
  \footnotesize
  \caption{Augmentation performance of different methods. $d_{1}\sim d_{3}$ are distances from different augmented domains to the source domain. $d_{intra}$ is the average pair-wise distance across augmented domains.}
    \begin{tabular}{c|cccc}
    \toprule
      Distance    & $d_{1}$   & $d_{2}$   & $d_{3}$   & $d_{intra}$ \\
    \midrule
    EERM  & 0.76  & 0.73  & 0.75  & 0.04 \\
    DPS-Random & 0.32  & 0.48  & 0.51  & 0.25 \\
    DPS   & 0.67  & 0.70   & 0.64  & 0.52 \\
    \bottomrule
    \end{tabular}%
  \label{distance}%
\end{wraptable}%
 It is noticeable that either removing $\mathcal{L}_{dist}$ or $\mathcal{L}_{kld}$ leads to a drop in performance. Hence, both $\mathcal{L}_{dist}$ and $\mathcal{L}_{kld}$ contribute to the performance of DPS.

Moreover, we study the augmentation performances of different methods. We compare the distance between the source domain and 3 augmented domain, denoted as $d_{1}\sim d_{3}$, and the average pair-wise distance across 3 augmented domains, denoted as $d_{intra}$. We employ the energy-score distance in OOD detection \cite{liu2020energy} as the distance metric. As shown in Table~\ref{distance}, the domains generated by EERM have small pair-wise distance, which shows that the augmented domains are very similar. DPS produces augmented domain with large intra-distance. Moreover, the augmented domains are far from the source domain. Hence, DPS can indeed introduce diverse OOD samples by generating diverse predictive subgraphs.



\section{Conclusion and Limitations}
\label{limitations}
In this work, we propose DPS to alleviate domain scarcity in graph domain generalization. DPS constructs multiple augmemted domains by finding diverse and predictable subgraphs from the source domains. The augmented domains are diverse and share the same semantics with the source domain, which avoid implausible augmentation. The generated domains facilitate learning a \textit{equi-predictive} GNN. 
DPS is model-agnostic that can be incorporated with various GNN backbones. 
Extensive experiments on both node-level and graph-level benchmarks shows the superior performance of DPS on various graph domain generalization tasks. The limitation is that we simplify the subgraph generation as the learnable node sampling process. We leave the improvement in our future work. 

\bibliographystyle{plain}
\bibliography{neurips_2022}

\end{document}